# A Model for Translation of Text from Indian Languages to Bharti Braille Characters

Nisheeth Joshi[1,2], Pragya Katyayan[1,2]
[1]Department of Computer Science, Banasthali Vidyapith, Rajasthan, India
[2]Centre for Artificial Intelligence, Banasthali Vidyapith, Rajasthan
nisheeth.joshi@rediffmail.com, pragya.katyayan@outlook.com

***Abstract*—People who are visually impaired face a lot of difficulties while studying. One of the major causes to this is lack of available text in Bharti Braille script. In this paper, we have suggested a scheme to convert text in major Indian languages into Bharti Braille. The system uses a hybrid approach where at first the text in Indian language is given to a rule based system and in case if there is any ambiguity then it is resolved by applying a LSTM based model. The developed model has also been tested and found to have produced near accurate results.**

***Keywords—Bharti Braille, Indian Languages, Transliteration, LSTM, Deep Learning***

## I. Introduction

Braille is a system of raised dots that can be felt with the fingertips and used to represent letters, numbers, and symbols. It was invented by Louis Braille, a French educator who was blind himself, in the early 19th century as a way for people who are blind or visually impaired to read and write. Braille consists of cells of six dots arranged in two columns of three dots each. Different combinations of dots represent different letters, numbers, and symbols. Braille is used worldwide as a standard writing system for people who are blind or visually impaired, and it is widely used for reading and writing in a variety of languages. The Braille system allows people who are blind or visually impaired to participate in activities such as reading, writing, and studying, which would otherwise be difficult or impossible for them, and it helps to break down barriers to information and education that are faced by this community.

Bharti Braille is a form of Braille script used in India to write the languages of India in Braille script. It is based on the standard Braille script but includes additional characters to represent the unique sounds and characters found in Indian languages.

The literary script is based on the standard Braille script but includes additional characters to represent the unique sounds and characters found in Indian languages. This includes characters for the retroflex sounds found in many Indian languages, as well as special characters to represent the unique conjunct consonants found in Indian languages.

Text to Braille Conversion is the process of converting written text into a system of raised dots called Braille, which can be felt with the fingertips and used by people who are blind or visually impaired to read. The Braille system consists of cells of six dots arranged in two columns of three dots each. Different combinations of dots represent different letters, numbers, and symbols. Text to Braille Conversion is typically done using specialized software or hardware devices. The software or devices can either be simple transliteration tools that follow a set of rules for mapping letters to Braille cells, or more advanced systems that use machine learning algorithms to improve the accuracy of the Braille representation.

Text to Braille Conversion systems are important because they provide a means for people who are blind or visually impaired to access written text and information. Braille is a tactile writing system that can be read by touch, and it allows people who are blind or visually impaired to participate in activities such as reading, writing, and studying, which would otherwise be difficult or impossible for them. Text to Braille Conversion systems make it possible to convert written text into Braille, so that it can be read by people who are blind or visually impaired, and they help to break down barriers to information and education that are faced by this community. In addition, Text to Braille Conversion systems can also be useful for proofreading Braille text, as well as for creating Braille materials for education, rehabilitation, and other purposes.

## II. Literature review

Hossain et al. [1] have identified rules and conventions for Bangla Braille translation based on rules. They proposed a DFA based computational model for MT, which gave acceptable translations. The results were tested by members of a visually impaired community.

Al-Salman et al. [2] have built a Braille copier machine which produced Braille documents in their respective languages. The machine worked as both copier as well as printing system using optical recognition and techniques from image processing. Yamaguchi et al. [3] have highlighted the problem of accuracy while translating technical notations to Braille. To solve this problem, they have developed a assistive solution for people from STEM background who are not capable of printing. Damit et al. [4] have mediated a new way of interlinking keyboard inputs from translations to commonly used Braille characters. This enabled visually blessed people to interact with visually impaired people.

Rupanagudi et al. [5] introduced a new technique of translating Kannada braille to Kannada language. They devised a new algorithm to segment Braille dots and identify characters. Choudhary et al. [6] have suggested a new approach for supporting communication amongst deaf-blind people. The technique included the use of smart gloves capable of translating Braille alphabets and can communicate the message via SMS. Due to this the user can convey simple messages using touch sensors. Guerra et al. [7] have developed a prototype using Java which can convert Braille text to digital text. Jariwala and Patel [8] have developed tool for translation of Gujarati, English and Hindi text to Braille and save it as a data file which can be directly printed via embosser.

Saxena et al. [9] have provided a real-time solution (hardware and software) for helping blind people. They developed a Braille hand glove which helped in communication for sending and receiving messages in real time. Nam and Min [10] have developed a music braille converted capable of converting the musical notations such as octaves, key signature, tie repeat, slur, time signature etc. successfully to Braille. Park et al. [11] have suggested a method of automatic translation of scanned images of books to digital Braille books. They implemented character identification and recognized images in Books while automatically translating them to text. This method reduced the time and cost required for producing books in Braille.

Alufaisan et al. [12] designed an application that identifies Braille numerals in Arabic and converts it to plain text using CNN-based Residual Networks. The system also gave speech assistance to the generated text. Apu et al. [13] proposed user and budget friendly braille device that can translate text and voice to braille for blind students. It works for different languages and converts based on 'text' or 'voice' command given by the user.

Yoo and Baek [14] have proposed a device that can print braille documents for blind. They implemented a raspberry Pi camera to save documents as images stored in device. Characters were extracted from the images and converted to Braille which is then processed to output braille. Their proposed device was portable and could be created using 3D printing. Zhang et al. [15] have used n-gram language model to implement Chinese-braille inter translation system. This system integrates Chinese and Braille word segmentation with concatenation rules. They have also proposed an experimental plan to improve Braille word segmentation amd concatenation rules with a word corpus of Chinese-Braille.

## III. CHALLENGES IN TRANSLATION OF TEXT INTO BHARTI BRAILLE

There are several issues that can arise during the conversion of text to Bharti Braille:

1. **Complex script structure:** The Devanagari script used in India has a complex structure, with multiple elements, including vowels, consonants, and diacritical marks, which can make the process of converting text to Bharti Braille challenging.

2. **Lack of standardization:** There is no standard set of rules for converting text to Bharti Braille, and different organizations and institutions may have their own variations in the mapping of Devanagari script characters to Braille cells.

3. **Incomplete or inaccurate transliteration rules:** If the rules for mapping Devanagari script characters to Braille cells are incomplete or inaccurate, the resulting Braille text may be incorrect or difficult to read.

4. **Technical limitations:** Text to Braille Conversion systems may have technical limitations, such as memory constraints, processing speed, or compatibility with certain software or hardware, which can affect the accuracy and efficiency of the conversion process.

5. **User training and awareness:** The success of Text to Braille Conversion systems also depends on user training and awareness. Users need to be trained on how to use the software or hardware correctly, and they need to be aware of the limitations and limitations of the system in order to use it effectively.

To overcome these issues, it is important to develop complete and accurate transliteration rules, to use software or hardware systems that are reliable and efficient, and to provide user training and awareness programs.

## IV. PROPOSED SYSTEM

To implement a translation system for Indian languages to Bharti Braille, we have used a hybrid approach which is a mix of rule-based and deep learning approaches. As a first step to our system, we have extracted phonemes from the input text and then created a rule base for mapping of Indian language-characters into Bharti Braille characters. Thus, we created rules for three characters classes viz. Consonants, Vowels and Vowel Symbols (matra/diacritics). The languages that we covered were: Hindi (Devnagari Script), Marathi (Devnagari Script), Nepali (Devnagari Script), Bengali (Bengali Script), Assamese ((Bengali Script), Gujarati (Gujarati Script), Punjabi (Gurumukhi Script), Odia (Odia Script), Tamil (Tamil Script), Telugu (Telugu Script), Kannada (Kannada Script), Malayalam (Malayalam Script), and Urdu (Perso-Arabic Script). Table 1 shows text in the respective languages and their Bharti Braille encoding.

TABLE I. INDIAN LANGUAGES TO BHARTI BRAILLE TRANSLATION

| Indian Language | Bharti Braille |
|---|---|
| আমার ভারত মহান | ⠜⠍⠜⠗ ⠘⠜⠗⠞ ⠍⠓⠜⠝ |
| મારું ભારત મહાન | ⠍⠜⠗⠥⠰ ⠘⠜⠗⠞ ⠍⠓⠜⠝ |
| मेरा भारत महान | ⠍⠑⠗⠜ ⠘⠜⠗⠞ ⠍⠓⠜⠝ |
| മേരാ ഭാരത് മഹാന് | ⠍⠑⠗⠜ ⠘⠜⠗⠞⠈ ⠍⠓⠜⠝⠈ |
| ମେରା ଭାରତ ମହାନ୍ | ⠍⠑⠗⠜ ⠘⠜⠗⠞ ⠍⠓⠜⠝⠈ |
| ਮੇਰਾ ਭਾਰਤ ਮਹਾਨ | ⠍⠑⠗⠜ ⠘⠜⠗⠞ ⠍⠓⠜⠝ |
| மேரா பாரத் மகான் | ⠍⠑⠗⠜ ⠏⠜⠗⠞⠰ ⠍⠅⠜⠰⠰ |
| మేరా భారత్ మహాన్ | ⠍⠑⠗⠜ ⠘⠜⠗⠞⠈ ⠍⠓⠜⠝⠈ |

The working of the system is shown in figure 1 where first we take input in an Indian language. Phonemes from the input text are extracted and based further the phonemes are divided into characters. These characters are mapped to Bharti Braille encodings using a rule base. If a particular character has more than two encodings, then the text is sent for disambiguation where a LSTM model tries to check the correct encoding based on the context.

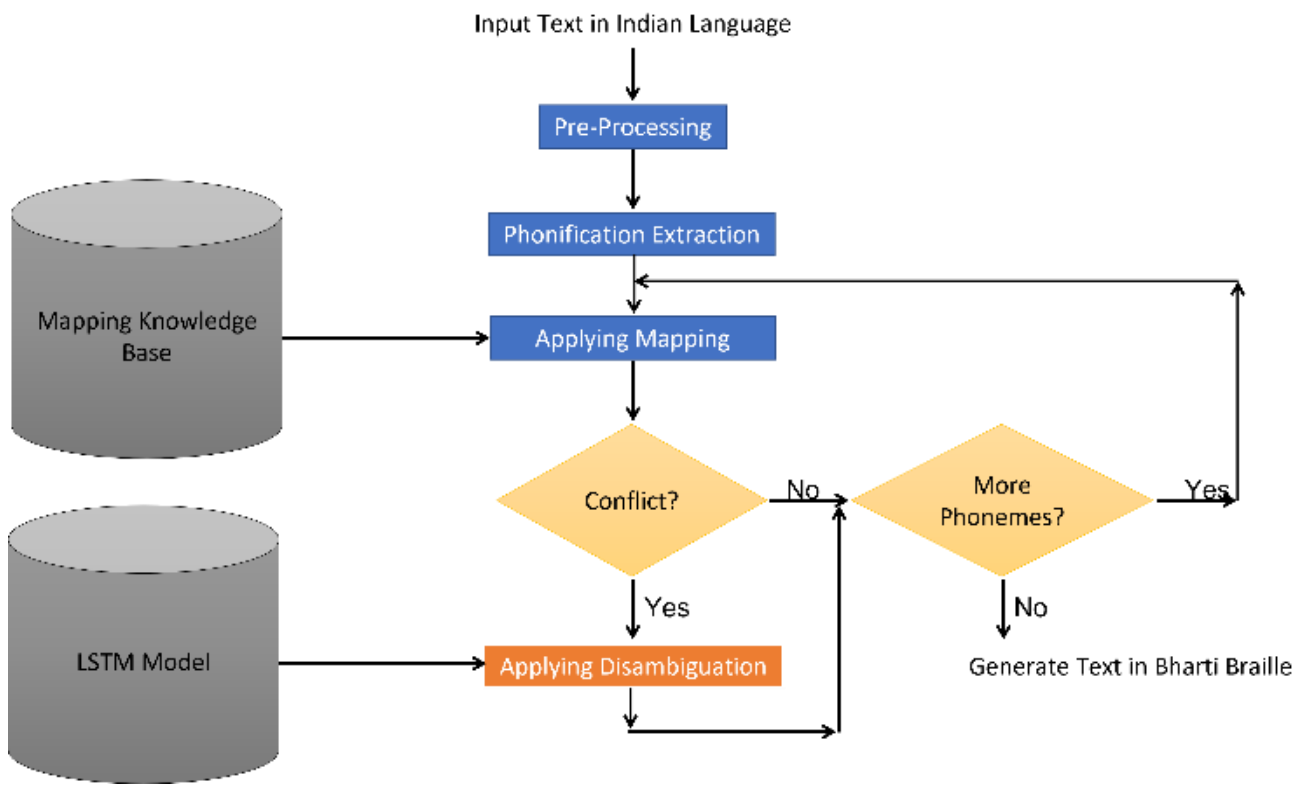

FIGURE 1: WORKING OF INDIAN LANGUAGES TO BHARTI BRAILLE TRANSLATION SYSTEM

The model takes in a sequence of tokens, $X=\{x1, x2,...,xT\}$, passes them through an embedding layer, $e$, to get the token embeddings, $e(X)=\{e(x1),e(x2),...,e(xT)\}$. These embeddings are processed - one per time-step - by the forward and backward LSTMs. The forward LSTM processes the sequence from left-to-right, while the backward LSTM processes the sequence right-to-left, i.e. the first input to the forward LSTM is $x$ and the first input to the backward LSTM is $xT$. The LSTMs also take in the hidden, $h$, and cell, c, states from the previous time-step.

$$h_t^{\rightarrow} = LSTM^{\rightarrow}(e(x_t^{\rightarrow}), h_{t-1}^{\rightarrow} c_{t-1}^{\rightarrow} \quad (1)$$

$$h_t^{\leftarrow} = LSTM^{\leftarrow}(e(x_t^{\leftarrow}), h_{t-1}^{\leftarrow} c_{t-1}^{\leftarrow} \quad (2)$$

After the whole sequence has been processed, the hidden and cell states are then passed to the next layer of the LSTM. The initial hidden and cell states, $h0$ and $c0$, for each direction and layer are initialized to a tensor full of zeros.

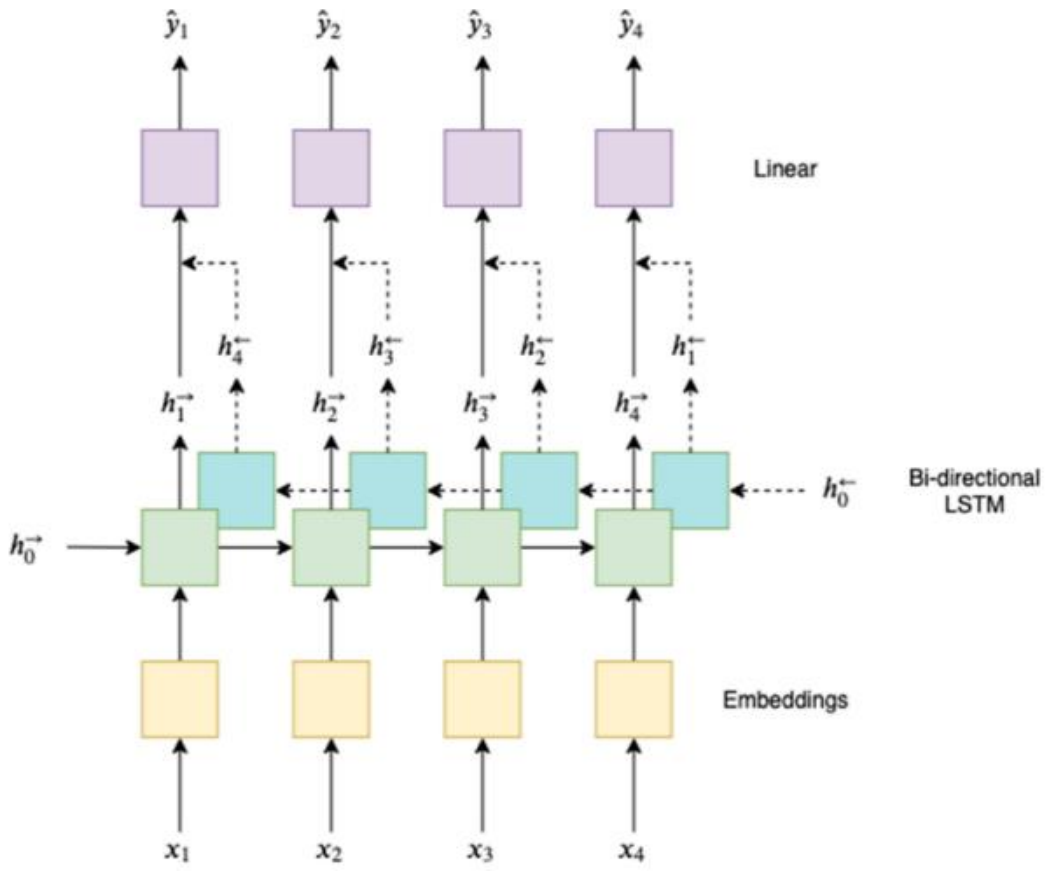

FIG. 2. WORKING OF BI-DIRECTIONAL LSTM LAYER WITH WORD EMBEDDINGS.

We then concatenate both the forward and backward hidden states from the final layer of the LSTM, $H=\{h1,h2,...hT\}$, where $h1=[h1^{\rightarrow}; hT'^{\rightarrow}]$, $h2=[h1'^{\rightarrow}; h(T1)'^{\rightarrow}]$, etc. and pass them through a linear layer, $f$, which is used to make the prediction of which tag applies to this token, $y'=f(ht)$.

When training the model, we compared our predicted tags, Y' against the actual tags, Y, to calculate a loss, the gradients with respect to that loss, and then updated our parameters. We also defined a dropout layer which we used in the forward method to apply dropout to the embeddings and the outputs of the final layer of the LSTM.

This is similar to what we have been doing in Markov chain in statistical learning theory, with the difference being that in statistical learning we apply probabilities and here we apply linear algebra to derive the values of previous, current and next states.

## V. EVALUATION

We tested our system for 1000 sentences for each Indian language. As per working shown in figure 1, rule-based system was applied first for of the input text and its Braille equivalent was obtained. In case if there was any ambiguity then the word having ambiguous character is sent to LSTM model which then resolves the conflict and provides with the final output. The systems were evaluated based on accuracy which is calculated based on equation 3. Across languages, the rule-based system gave varied accuracy and LSTM gave 100% accuracy, thus resulting in 100% overall accuracy. Table 2 shows the results of this evaluation study.

$$Accuracy = \frac{Correct\ Mappings}{Total\ Characters} \quad (3)$$

TABLE II. SUMMARY OF EVALAUTION

| Script | Rule Based | LSTM | Total |
|---|---|---|---|
| **Devanagari** | 0.94 | 1 | 1 |
| **Bengali** | 0.938 | 1 | 1 |
| **Gujarati** | 0.939 | 1 | 1 |
| **Kannada** | 0.9361 | 1 | 1 |
| **Malayalam** | 0.9423 | 1 | 1 |
| **Odia** | 0.934 | 1 | 1 |
| **Punjabi** | 0.9398 | 1 | 1 |
| **Tamil** | 0.925 | 1 | 1 |
| **Telugu** | 0.928 | 1 | 1 |
| **Urdu** | 0.938 | 1 | 1 |

## VI. CONCLUSION

In this paper we have shown the development of translation system from Indian languages to Bharti Braille. For this we developed the transcription system for 13 languages in 10 scripts. A two-pronged hybrid mechanism was used were a rule-based system performed the first step of translation and in case of any ambiguities an LSTM model was use for resolution and correct translation into Bharti Braille. Overall, the system produced near 100% accurate results across languages.

## ACKNOWLEDGMENT

This work is supported by the funding received from SERB, GoI through grant number CRG/2020/004246 for project entitled, "Development of English to Bharti Braille Machine Assisted Translation System".